\documentclass[10pt,twoside]{amundi_article}

\usepackage[table]{xcolor}
\usepackage{lmodern}
\usepackage{amssymb}
\usepackage{amsfonts}
\usepackage{amsmath}
\usepackage{amsbsy}
\usepackage{fancybox}
\usepackage{fancyhdr}
\usepackage{graphicx}
\usepackage[hyphens]{url}
\usepackage{arydshln}
\usepackage{multirow}
\usepackage{pdflscape}
\usepackage{tikz}
\usepackage{comment}
\usepackage{nicefrac}
\usepackage{dsfont}
\usepackage{algorithmic}
\usepackage{algorithm}
\usepackage{lipsum}

\newlength{\figurewidth}
\newlength{\figureheight}
\DeclareMathAlphabet\mathbfcal{OMS}{cmsy}{b}{n}
\DeclareFontFamily{OT1}{pzc}{}
\DeclareFontShape{OT1}{pzc}{m}{it}{<-> s * [1.3] pzcmi7t}{}
\DeclareMathAlphabet{\mathpzc}{OT1}{pzc}{m}{it}

\usetikzlibrary{positioning}
\usetikzlibrary{patterns,arrows,decorations.pathreplacing}
\usetikzlibrary{backgrounds}

\usepackage[
        left=3.0cm,
        right=3.0cm,
]{geometry}

\newlength\FHoffset
\fancyheadoffset{\FHoffset}

\usepackage[authoryear]{natbib}

\usepackage{ragged2e}

\usepackage{subfigure}
\usepackage{booktabs}
\definecolor{amundi_blue}{RGB}{0,176,240}
\definecolor{amundi_dark_blue}{RGB}{0,28,75}
\definecolor{darkblue}{rgb}{0.0, 0.0, 0.55}
\usepackage{hyperref}
\hypersetup{
    colorlinks=true,
    citecolor=darkblue,
    linkcolor=darkblue,
    filecolor=darkblue,      
    urlcolor=darkblue
}

\usepackage{pdfpages}
\usepackage{bookmark}

\usepackage{enumitem}
\usepackage{balance}

\DeclareMathOperator*{\argmax}{arg\,max}

\usepackage{tabularx}

\begin{document}

\setcounter{page}{1}

\title{\textbf{\color{amundi_blue}A Novel Classification Approach for Credit Scoring based on Gaussian Mixture Models
}%
}

\author{
\hspace{-0.3cm}
{\color{amundi_dark_blue} Hamidreza Arian}\footnote{hamidreza.arian@utoronto.ca} \\
\hspace{-0.3cm} Sharif University \\
\hspace{-0.3cm} of Technology
\and
{\color{amundi_dark_blue} Seyed Mohammad Sina Seyfi}\footnote{seyfi.sina@aut.ac.ir} \\
Khatam \\
University
\and
{\color{amundi_dark_blue} Azin Sharifi}\footnote{azin.sharifi@utoronto.ca} \\
Sharif University \\
of Technology
}

\date{\color{amundi_dark_blue} August 2019}

\maketitle

\begin{abstract}
\noindent
Credit scoring is a rapidly expanding analytical technique used by banks and other financial institutions. Academic studies on credit scoring provide a range of classification techniques used to differentiate between good and bad borrowers. The main contribution of this paper is to introduce a new method for credit scoring based on Gaussian Mixture Models. Our algorithm classifies consumers into groups which are labeled as positive or negative. Labels are estimated according to the probability associated with each class. We apply our model with real world databases from Australia, Japan, and Germany. Numerical results show that not only our model's performance is comparable to others, but also its flexibility avoids over-fitting even in the absence of standard cross validation techniques. The framework developed by this paper can provide a computationally efficient and powerful tool for assessment of consumer default risk in related financial institutions.  
\end{abstract}



\section{Introduction}\label{sec:intro}

Credit scorecard models have gained considerable attention from both academics and practitioners in the last two decades (\cite{baesens2003benchmarking,lessmann2015benchmarking,brown2012experimental}). According to~\cite{thomas2002credit}, 
credit scoring is a set of decision making models which helps financial institutions in their credit lending operations. Current statistical methods provide significant contributions to the field of machine learning. These techniques are employed for building modern models to measure the risk level of a single customer based on their characteristics. The resulting model helps risk managers classify customers as good or bad applicants according to their risk level. Therefore, the main application of credit scoring models is to recognize the features that impact the payment behavior of customers as well as their default risk (\cite{han2006data}). 

A comprehensive literature review was given in the study of \cite{hand1997statistical}, which discusses important classification methods applied to credit scoring. They review several statistical approaches and proposed that the future trend of credit scoring models will be more novel and sophisticated. For indicating studies of modern methods, \cite{lee2005two} introduce a discriminant neural network to perform credit rating.  \cite{van2006bayesian}, propose a support vector machine model
within a Bayesian evidence framework. \cite{hoffmann2002comparing}, study a boosted genetic fuzzy model, \cite{hsieh2010data}, analyze a combined method using neural networks, support vector machine and Bayesian networks. 
Among these classic papers, \cite{west2000neural} compared five neural network models with traditional techniques. The results indicate that neural networks can improve the credit scoring accuracy. \cite{baesens2003benchmarking} perform a comparison involving discriminant analysis, logistic regression, logic programming, support vector machines, neural networks, Bayesian networks and k-nearest neighbor. The authors conclude that many classification techniques give performances which are quite competitive with each other. 

In order to improve the accuracy of credit scoring models, address data imbalance problems, and eliminate over-fitting, extensive research has been done on development of hybrid and combined methods using various above mentioned techniques, including Support Vector Machines (\cite{chen2010combination}), Decision Trees (\cite{chen2010combination,zhang2010vertical}), Fuzzy Classifiers 
with Boosting (\cite{hoffmann2002comparing}), Ensemble Classification techniques (\cite{nanni2009experimental,paleologo2010subagging,xiao2016ensemble}), Neural Networks and Multi-variate Adaptive Regression Splines (\cite{hsieh2005hybrid,lee2005two,hsieh2010data}). These new techniques are shown to improve model simplicity, model speed, and model accuracy (\cite{liu2005data,martens2007comprehensible}). In addition, \cite{ghodselahi2011hybrid} provides a comparative analysis on ten SVM classifiers as the members of the ensemble model and show the SVM classifier works as well as decision tree classifier. They also represented a hybrid GASVM model by combining genetic algorithms with SVM classifier. Support Vector Machines are also capable of being a hybrid component with data mining techniques, where clustering and classification stages are done through SVM (\cite{chen2012credit}). Additionally, Cluster Support Vector Machine (CSVM) for credit scoring has been reported as an efficient classifier compared to other nonlinear SVM methods (\cite{harris2015credit}). There are contrasting results, however, on efficiency of combination of multiple classifiers for credit scoring (\cite{tsai2008using,wang2011comparative}). These comparative studies have included a variety of classification techniques to formulate the credit scoring problem. In the group of these current scorecard models, each have their own strengths and weaknesses. Therefore, there seem to be no clear best credit scoring technique. 

In this paper, we will derive our novel clustering algorithm based on the application of Gaussian Mixture Models in credit scoring. Gaussian Mixture Models (GMMs), are one of the most widely used unsupervised mixture models for clustering (\cite{bishop2006pattern}). Each Gaussian density is called a component of the mixture and has its own mean and covariance. In many GMM applications, parameters are determined by maximum likelihood, typically using the Expectation-Maximization algorithm (\cite{dempster1977maximum}). GMM has been commonly used in a wide range of applications such as data mining, pattern recognition, as well as signal and image processing, but so far, their application in building scorecard models is missing in the literature. 

There are a series of informative surveys covering various types of credit scoring techniques. First, \cite{baesens2003benchmarking} provide a benchmarking research updated after ten years by \cite{lessmann2015benchmarking} in which a comparison between 41 classifiers and eight algorithms is studied. Two other examples, \cite{hand1997statistical} and \cite{thomas2000survey} provide reviews on statistical and operational research methods for consumer credit classification. Finally, a very stimulating paper by \cite{louzada2016classification} present an interesting survey of 184 articles in 73 journals in credit scoring, and quite surprisingly Gaussian Mixture Models are not studied in any of them. However, there are some studies to address the problem of imbalance data arising from higher number of healthy borrowers in most credit scoring databases. For instance, \cite{han2019gaussian} employs Gaussian Mixture Models for data re-sampling to eliminate imbalance data problem in consumer credit market. 

The rest of this paper is structured as follows. In Section \ref{sec:method}, we first briefly introduce credit features and Gaussian Mixture models, and then present our conceptual Credit Scoring Gaussian Mixture (termed as CSGM) model, with a detailed description of its clustering, classification and assignment steps. Section \ref{sec:emp} gives a comparative analysis of our approach with five other standard models. A managerial guideline and applications of our model for financial institutions and commercial banks are discussed in Section \ref{sec:business} and concluding remarks in Section \ref{sec:con} end the paper. 
\section{Methodology}\label{sec:method}

We first provide a brief overview of standard credit features, and then use GMMs to build a credit scoring model.

\subsection{Credit Features}

Assume the data set consists of N samples $\mathbf{x_1},\mathbf{x_2},...,\mathbf{x_N}$, where $\mathbf{x_i}=(x_1^i, x_2^i, \cdots, x_d^i)$, for $i = 1,2, \cdots ,N$ is a d-dimensional vector. Each data point $\mathbf{x_i}$ can belong to a discrete set of samples shown by $\{y_i\}_1^m$ with $m$ being the set of credit score classes. Here in this article, we have two possible credit classes, namely $y_i\in {0,1}$, but emphasize that credit scoring systems can have more than two states. For instance, a scorecards model can separate positive scores into individuals for whom their existing loan can be extended in near future or not. 
The data set contains two subsets $S^+$ and $S^-$. Samples in $S^+$ share the same class label $y=1$, and those in $S^-$ have class label $y=0$. 

In terms of the clustering structure, we assume there are $N_c$ clusters $C_1, C_2, ..., C_{N_c}$. Centroids of clusters are $\boldsymbol{\mu_1},$ $\boldsymbol{\mu_2}$, $...$, $\boldsymbol{\mu_{N_c}}$, and the covariances of the Gaussian Mixture components are $\boldsymbol{\Sigma_1},$ $\boldsymbol{\Sigma_2}$, $...$, $\boldsymbol{\Sigma_{N_c}}$, represented by  
\begin{equation}
    \boldsymbol{\mu_k}=\frac{1}{n_k}\sum_{j \in I_k}{\mathbf{x_j}} \quad \text{and} \quad 
    \boldsymbol{\Sigma_k} = \frac{1}{n_k}\sum_{j \in I_k}{(\mathbf{x_j}-\boldsymbol{\mu_k})(\mathbf{x_j}-\boldsymbol{\mu_k})^T},
\end{equation}
where $I_k$ is the index set of all samples in the $k$th cluster, $n_k=|I_k|$ is the number of data samples in cluster $C_k$ and $\mathbf{x_j}$, with $j$ being in the index set $I_k$, denotes a sample that belongs to the cluster $C_k$.

\subsection{Clustering with Gaussian Mixture Models}

In mathematical terms, a GMM is a parametric probability distribution written as a weighted sum of $N_c$ component Gaussian distributions
\begin{equation}
p(\mathbf{x} | \boldsymbol{\theta})=\sum_{i=1}^{N_c} w_{i} \phi\left(\mathbf{x} | \boldsymbol{\mu_{i}}, \boldsymbol{\Sigma}_{i}\right) , 
\end{equation}
where $\mathbf{x}$ is a $d$-dimensional feature vector, the triplet $\boldsymbol{\theta}=(\boldsymbol{\omega}, \boldsymbol{\mu}, \boldsymbol{\Sigma})$ gives the parameters of the model, $\omega_i$s are the weights summing up to one, and $\phi\left(\mathbf{x} | \boldsymbol{\mu_{i}}, \mathbf{{\Sigma}_{i}}\right)$ represents the individual Gaussian density with mean $\boldsymbol{\mu_i}$ and co-variance $\boldsymbol{\Sigma_i}$ 
\begin{equation}
{\phi\left(\mathbf{x} | \boldsymbol{\mu_{i}}, \boldsymbol{\Sigma_{i}}\right)=\frac{1}{(2 \pi)^{D / 2}\left|\boldsymbol{\Sigma_{i}}\right|^{1 / 2}}} { \exp \left\{-\frac{1}{2}\left(\mathbf{x}-\boldsymbol{\mu_{i}}\right)^{\prime} \boldsymbol{\Sigma_{i}}^{-1}\left(\mathbf{x}-\boldsymbol{\mu_{i}}\right)\right\}}.
\end{equation}
In the Gaussian Mixture Model, a $N_c$-dimensional latent variable $\mathbf{z}$ is defined according to a $1$-of-$N_c$ representation $\mathbf{z} = (z_1, z_2, \allowbreak \cdots, z_{N_c})$ with $z_i\in \{0, 1\}$ and $\sum_i{z_i}=1$. Then the joint distribution is $p(\mathbf{x},\mathbf{z})=p(\mathbf{z})p(\mathbf{x}|\mathbf{z})$, where the conditional probability component can be written as 
\begin{equation}
\label{eq5}
p(\mathbf{x}|z_k=1)=\phi(\mathbf{x}|\boldsymbol{\mu_k},\boldsymbol{\Sigma_k}) .
\end{equation}
 
Gaussian Mixture Models provide an unsupervised clustering approach as they are blind to the label of input data. In our CSGM model, we do not make assumptions on the creditworthiness of each individual up to the point when the clusters are formed. We give the classification labels to each cluster after the GMM model with the optimal number of components is calibrated.

\subsection{Parameter Estimation}

During the learning process in any scorecard algorithm, model parameters should be estimated. First, assume there are $N$ credit scoring feature vectors in our learning set
\begin{equation}
\mathbf{x} = \left\{\mathbf{x_1}, \mathbf{x_2}, \cdots, \mathbf{x_N}\right\} .
\end{equation}
Since the expression of the GMM likelihood function is a non-linear function of its parameter set $\boldsymbol{\theta}$, standard methods do not work. We use the Expectation-Maximization algorithm (EM) to find an appropriate $\boldsymbol{\theta}$ which maximizes the Likelihood function
\begin{equation}
    \boldsymbol{\theta_{ML}} = \underset{\boldsymbol{\theta}}{\argmax}(\log p(\mathbf{x}|\boldsymbol{\theta})) .
\end{equation}
The EM algorithm is a highly efficient iterative technique to find maximum likelihood solutions for Gaussian Mixture Models\footnote{see \cite{dempster1977maximum}}. EM algorithm iteratively moves in the domain of model parameters, $\mathcal{D}$, from one step's solution $\boldsymbol{\theta}_n \in \mathcal{D}$ to the next $\boldsymbol{\theta}_{n+1} \ \in \mathcal{D}$ in such a way that the likelihood function is monotonically increasing, i.e. $p(\mathbf{x}|\boldsymbol{\theta}_n) \geq p(\mathbf{x}|\boldsymbol{\theta_{n+1}})$.

In the EM algorithm, we first initialize the parameter vector $\boldsymbol{\theta_0}$ by a possible set of weight, mean and covariance components. Then in the Expectation step, using previous parameters of the algorithm, we compute the responsibility functions
\begin{equation}
r\left(z_{n k}\right)=\frac{\omega_{k} \phi\left(\mathbf{{x}_{n}} | \boldsymbol{\mu_{k}}, \boldsymbol{\Sigma_{k}}\right)}{\sum_{j=1}^{N_c} \omega_{j} \phi\left(\mathbf{{x}_{n}} | \mathbf{\boldsymbol{\mu_j}}, \boldsymbol{\Sigma_j}\right)} . \label{probfunc}
\end{equation}
Next in the Maximization step, we update model parameters by
\begin{equation}
\begin{aligned} \boldsymbol{{\mu}_{k}^{\text { new }}} &=\frac{1}{n_k} \sum_{n=1}^{N} r\left(z_{n k}\right) \mathbf{{x}_{n}}, \quad {\omega}_{k}^{\text { new }} =\frac{\sum_{n=1}^{N} r\left(z_{n k}\right)}{N}\\
\mathbf{{\Sigma}_{k}^{\text { new }}} &=\frac{1}{n_k} \sum_{n=1}^{N} r\left(z_{n k}\right)\left(\mathbf{{x}_{n}}-\boldsymbol{{\mu}_{k}^{\text { new }}}\right)\left(\mathbf{{x}_{n}}-\boldsymbol{{\mu}_{k}^{\text { new }}}\right)^{\mathrm{T}} .
\end{aligned}
\end{equation}
The termination criteria is set by when improvements of the likelihood function stop.

\subsection{Estimating Number of GMM Components}

We use model selection methods based on information criteria to determine the number of components per GMM to best capture data distribution. Information criteria provide a technique for maximizing training data's likelihood while trying to avoid over-fitting. The Akaike Information Criterion (AIC)\footnote{see~\cite{akaike1974new}} is a model selection method, formally defined as
\begin{equation}
\text{AIC}(\boldsymbol{\theta}) = 2 k - 2 \log p(\mathbf{x} | \boldsymbol{\theta}) ,
\end{equation}
where $\boldsymbol{\theta}$ represents the model parameter set, $\mathbf{x}$ is the observed data, and $k$ is the number of parameters estimated by the model. We also use the Bayesian Information Criterion (BIC)\footnote{see~\cite{schwarz1978estimating}} 
\begin{equation}
\text{BIC}(\boldsymbol{\theta}) = k \log (N) - 2 \log p(\mathbf{x} | \boldsymbol{\theta}) ,
\end{equation}
which penalizes model complexity more heavily than AIC. Here $N$ is the number of data points. The model selected will have the lowest BIC or AIC scores.

\subsection{Binary Classification Using GMM} \label{sec:methodprob}

Here we provide an algorithm for label assignment to both GMM generated clusters and individual data. We perform this task in the four steps discussed below. Our objective is to classify data into two groups with either \textit{good} or \textit{bad} credit. 

\begin{enumerate}[label=\bfseries\arabic*,leftmargin=*,align=left]
\item[\textbf{2.5.1}] \textbf{Associating individual data to clusters:} 
We allocate individual data to $N_c$ different clusters by calculating the probability $p(\mathbf{x_i}\in z_k)$
\begin{equation} \label{probgmm22}
p(\mathbf{x_i} \in z_k)=\frac{\omega_{k} \phi\left(\mathbf{x_i}| \boldsymbol{{\mu}_{k}}, \boldsymbol{\Sigma}_{k}\right)}{\sum_{j=1}^{N_c} \omega_{j} \phi\left(\mathbf{x_i}| \boldsymbol{\mu}_{j}, \boldsymbol{\Sigma}_{j}\right)} . 
\end{equation}
We call $p(\mathbf{x_i}\in z_i)$, \textit{dependency probability} of data $\mathbf{x_i}$ to cluster $z_i$. Each individual data $\mathbf{x_i}$ belongs to the cluster $z_i$ with highest probability $p(\mathbf{x_i}\in z_i)$. Therefore, $\mathbf{x_i}$ belongs to cluster $k$ if 
\begin{equation}
k = \underset{k}{\argmax}(p(\mathbf{x_i} \in z_k)).
\end{equation}
In the improbable case of multiple $j$s satisfying the right hand side equation, we pick one arbitrarily. 

\item[\textbf{2.5.2}] \textbf{Labeling clusters based on the associated train data:} We associate $r^*(1 | z_k)$ to cluster $k$ by counting the number of its members with good credit and then divide it by the total number of class members $n_k$. We define the ratio $r^*(0 | z_k)$ similarly such that $r^*(0 | z_k) + r^*(1 | z_k) = 1$. Next, each cluster is labeled such that
\begin{equation}\label{eq:labeling_cluster}
    l_{z_k}=\begin{cases}
			1, & r^*(1|z_k) > r^*(0|z_k)\\
            0, & \qquad \text{otherwise}
		 \end{cases} .
\end{equation}
\item[\textbf{2.5.3}] \textbf{Setting the decision boundary:} After labeling clusters, we define a decision boundary $0 \leq D \leq 1$ by which an arbitrary training data $\mathbf{x_i}$ can be labeled either $0$ or $1$. Now we calculate the probability 
\begin{equation} \label{eq:probabilityratio}
    p(y = 1|\mathbf{x_i}) = \sum_{k=1}^{N_c}{r^*(1|z_k) p(\mathbf{x_i}\in z_k)}  , 
\end{equation}
where $y$ is the target. The same probability can be calculated for $p(y = 0|\mathbf{x_i})$ using $r^*(0|z_k)$ and clearly $p(y = 0|\mathbf{x_i}) + p(y = 1|\mathbf{x_i}) = 1$. 

\item[\textbf{2.5.4}] \textbf{Labeling train data based on cluster's label:} We are now able to label each individual training data based on the choice of decision boundary $D$ and the probability $p(y = 1|\mathbf{x_i})$
\begin{equation}\label{eq:labeling_data}
l_{\mathbf{x_i}}=
\begin{cases}
    1, & p(y = 1|\mathbf{x_i}) > D\\
    0, & \quad \text{otherwise}
\end{cases} . 
\end{equation}

\end{enumerate}

The above steps provide our binary CSGM classification algorithm for labeling training data. For labeling an individual test data $\mathbf{x}$, we fix parameters of the second and third step including $r^*$ ratios for clusters and decision boundary $D$. Next, we run steps 1 and 4 to calculate the probabilities $p(y = 1|\mathbf{x})$ and make decision on the label of $\mathbf{x}$ by boundary $D$. In Section~\ref{sec:business}, we discuss how the decision boundary $D$ provides the flexibility to risk managers for making a trade-off between precision and recall.
\begin{figure*}
\centering
\subfigure[Australia]{{\includegraphics[width=4.7cm]{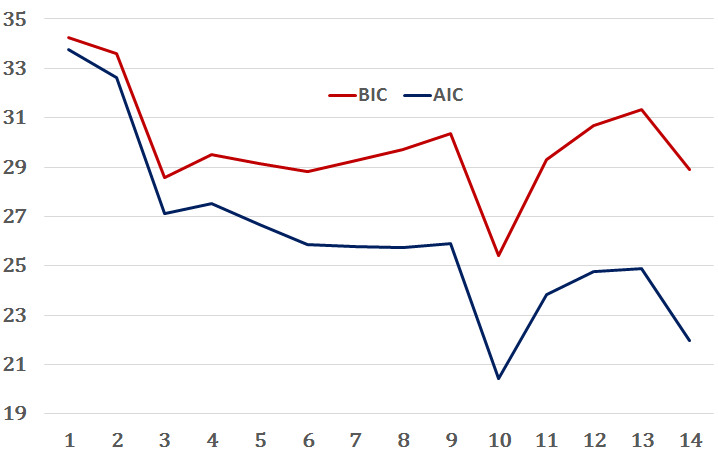}}}
\quad
\subfigure[Japan]{{\includegraphics[width=4.7cm]{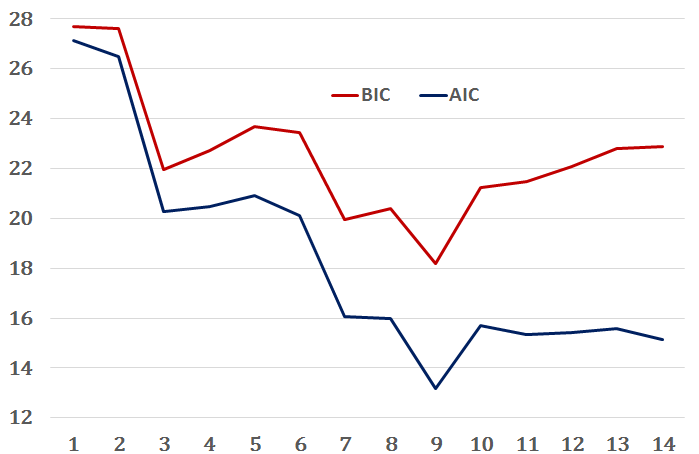} }}%
\quad
\subfigure[German]{{\includegraphics[width=4.7cm]{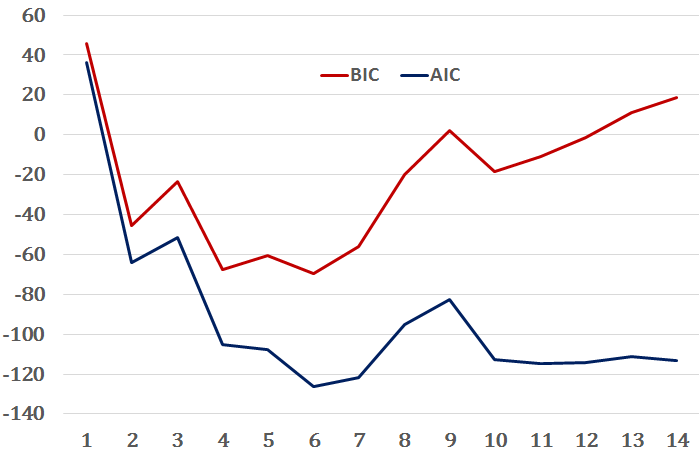} }}\\
\caption{AIC and BIC measures for optimal number of components (horizontal axis: the number of Gaussian components, vertical axis: the information criteria (in thousands)).}
\label{AICBICall}%
\end{figure*}

\section{Empirical Analysis and Results}\label{sec:emp}

Here we give numerical results of our model on credit scoring data sets from Germany, Australia and Japan. We compare the accuracy of our model with some of the existing models in the literature, using our Python implementation.\footnote{If you would like to see our Python implementation, please contact the corresponding author. After our article is accepted, we plan to publish our code on Github.} 

\subsection{Real Credit Scoring Data}

We test our CSGM model on three real world credit data sets from UCI Machine Learning Repository\footnote{\cite{Dua:2019}}: (1) German Credit Scoring, (2) Australian Credit Approval and (3) Japanese Credit Screening. General information on the three data sets can be seen in table~\ref{tab:introdata}.

\begin{table}
  \centering
  \caption{Summary of German, Australian and Japanese Data Sets}
\begin{tabularx}{\textwidth}{XXXXXX}
    \toprule
     Country &  Num. of & Num. of  & Healthy & Categorical & Numerical \\
      & Classes & Samples & Data & Features & Features \\
     \midrule
    Germany & 2     & 1000  & 700   & 13    & 7      \\
    Australia & 2     & 690   & 307   & 5     & 9     \\
    Japan & 2     & 661   & 297   & 10    & 5    \\
    \bottomrule
    \end{tabularx}%
  \label{tab:introdata}%
\end{table}%

The German credit data set consists of 1000 samples with 20 attributes; 7 numerical and 13 categorical. For the German data, there are 700 samples with good credit and 300 samples with bad credit. We have converted the categorical features to dummy variables which increased the number of features for the German data set to 62.

The Australian Credit Approval, includes 690 instances, with 14 attributes; 6 numerical and 8 categorical. We replace categorical features with integers for the sake of statistical convenience. For the Australian data set, there are 383 samples labeled 0 and 307 samples labeled 1. Here, 0 and 1 indicate bad and good credit scores, respectively. 

The Japanese data set, includes 15 features; 5 numerical and 10 categorical. Here again we substitute symbols by integers for convenience. We also eliminate all samples with missed values in Japanese data set, resulting in a decrease in the number of samples from 690 to 661. 


\subsection{Experimental Results}

We use the information criteria BIC and AIC for selecting the number of GMM components, where these measures are minimized. This results in $N_c$ components, to which all input samples can be allocated. 

\begin{table}
  \centering
  \caption{Train and Test Accuracy of Various Clusters}
  \begin{tabularx}{\textwidth}{XXXXXXXXX}
\toprule
No.   & \multicolumn{2}{c}{German} & No.   & \multicolumn{2}{c}{Australia} & No.   & \multicolumn{2}{c}{Japan} \\
Classes & train & test  & Classes & train & test  & Classes & train & test \\
\midrule
1     & 63\%  & -     & 1     & 70\%  & 78\%  & 1     & 76\%  & 81\% \\
2     & 73\%  & 73\%  & 2     & 100\% & -     & 2     & 100\% & - \\
3     & 99\%  & 47\%  & 3     & 83\%  & -     & 3     & 100\% & - \\
4     & 74\%  & -     & 4     & 100\% & -     & 4     & 100\% & - \\
5     & 62\%  & 20\%  & 5     & 100\% & -     & 5     & 100\% & - \\
6     & 67\%  & -     & 6     & 60\%  & 25\%  & 6     & 90\%  & 92\% \\
      &       &       & 7     & 78\%  & 80\%  & 7     & 73\%  & 85\% \\
      &       &       & 8     & 100\% & -     & 8     & 100\% & - \\
      &       &       & 9     & 100\% & -     & 9     & 87\%  & 71\% \\
      &       &       & 10    & 95\%  & 95\%  &       &       &  \\
\bottomrule
    \end{tabularx}%
  \label{class_acc}%
\end{table}%

Following our modeling approach in section~\ref{sec:methodprob}, during the training period, each of $N_c$ clusters will be labeled 0 and 1. The clusters and their trained parameters are then used to associate a cluster to each test sample. The model accuracy is calculated by ratio of successful predictions to the total number of samples. The covariances of GMM's components are assumed to be full-ranked matrices.

We split each data set to train and test by a $2$-to-$1$ ratio. The German data set is imbalanced with $70\%$ good credit applicants. Imbalanced data sets have a majority class with a large number of members scrambled with a minority class with much fewer members. For handling such data sets, we use Synthetic Minority Oversampling Technique (SMOTE)\footnote{\cite{chawla2002smote}}. In SMOTE, an arbitrary minority member, $\mathbf{x_1}$, with its nearest neighbour, $\mathbf{x_2}$, in the same class are picked. Assume the vector from the first to the second member is $\mathbf{x_{1 \to 2}} = \mathbf{x_2} - \mathbf{x_1}$. Then a uniform random variable $0 < \alpha < 1$ is generated and a third member $\mathbf{x_3} = \mathbf{x_1} + \alpha \mathbf{x_{1 \to 2}}$ is added to the minority class. We repeat this process enough times until the data set is balanced. For the case of German data set, before SMOTE, the majority and minority classes had 458 and 208 members, respectively. After SMOTE both classes contain 458 data points.

\begin{figure*}
\centering
\subfigure[German Train set]{{\includegraphics[width=5.0cm]{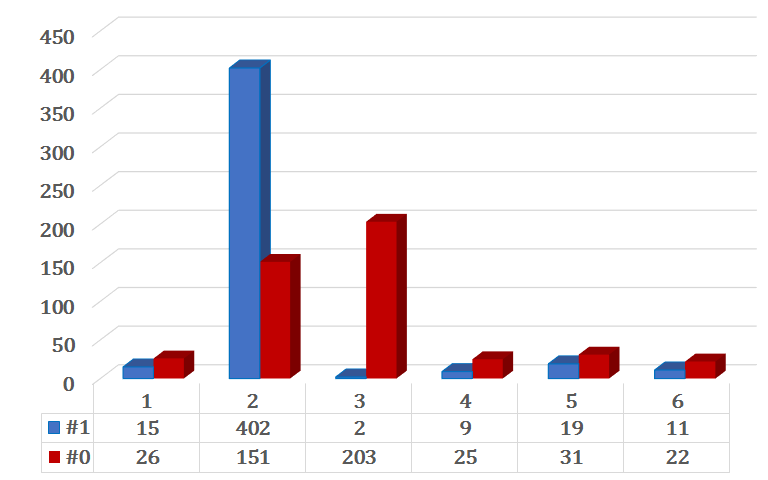}}}
\subfigure[Australia Train set]{{\includegraphics[width=5.0cm]{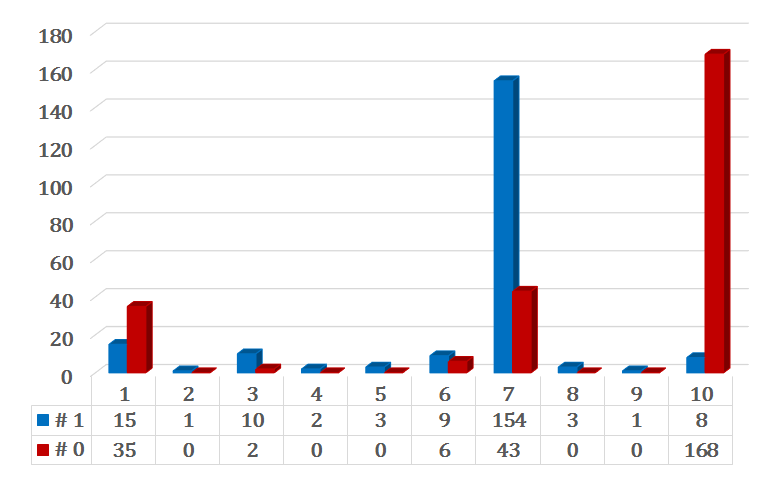}}}
\subfigure[Japan Train set]{{\includegraphics[width=5.0cm]{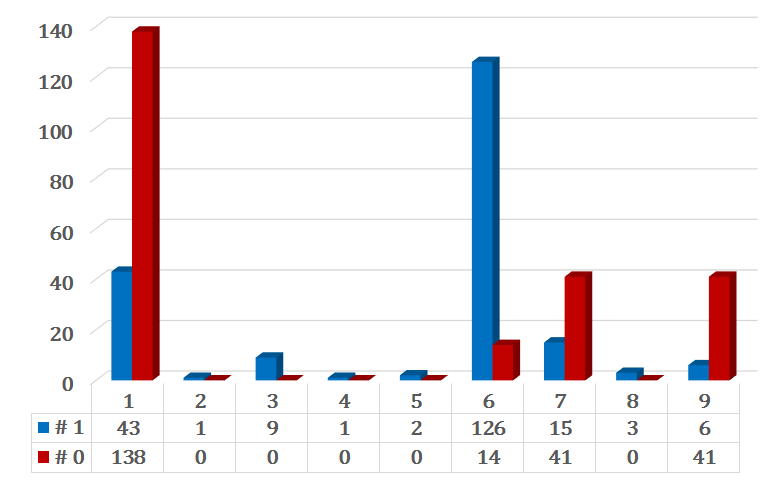}}}\\
\subfigure[German Test Set]{{\includegraphics[width=5.0cm]{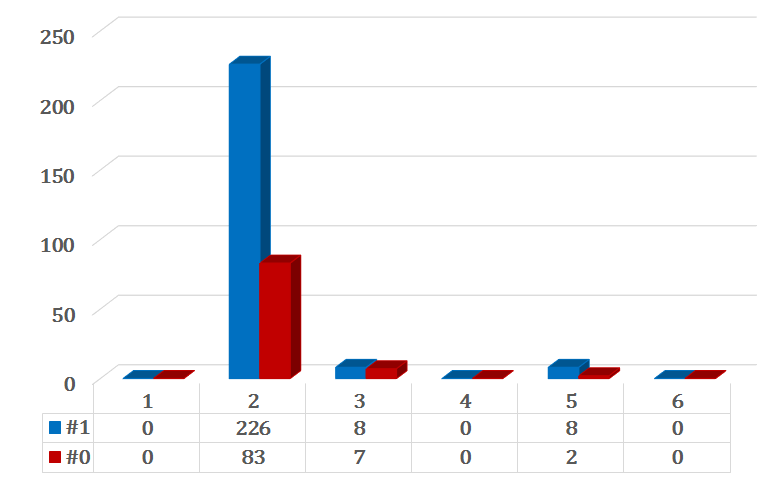} }}
\subfigure[Australia Test Set]{{\includegraphics[width=5.0cm]{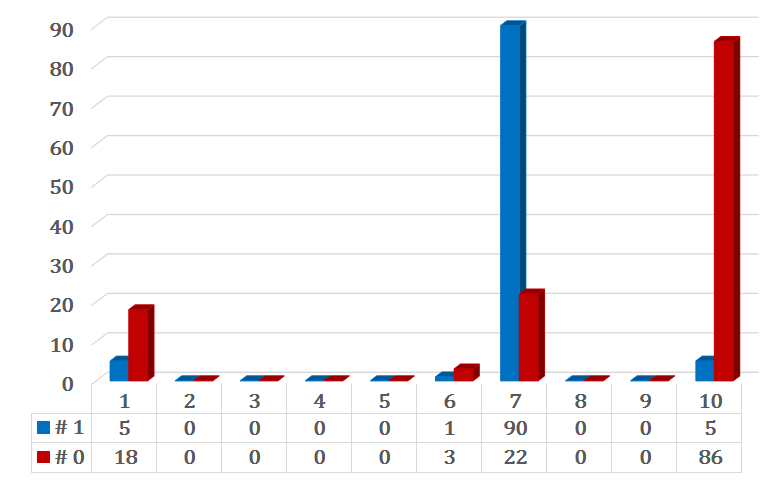} }}
\subfigure[Japan Test Set]{{\includegraphics[width=5.0cm]{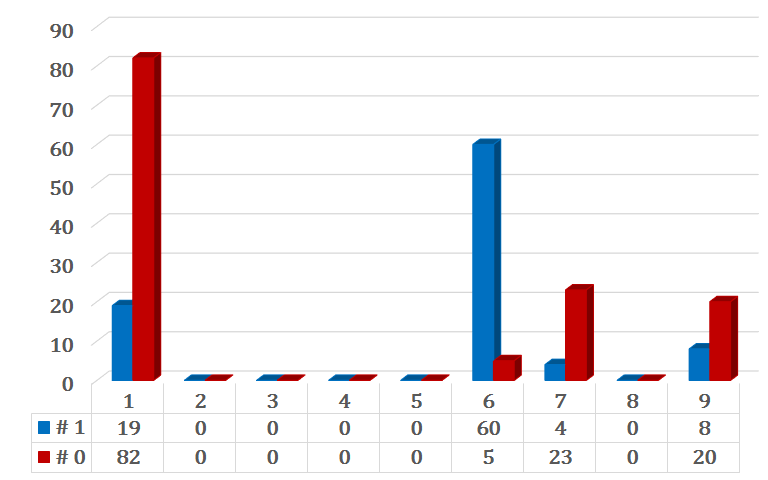} }}%
\caption{Number of 1s and 0s in train and test sets for Germany (panels (a) and (d)), Australia (panels (b) and (e)), and Japan (panels (c) and (f))}
\label{fig:labeling_clusters}
\end{figure*}

In figure \ref{AICBICall}, we plot AIC and BIC measures which give the optimal number of components, $N_c$, for each of the three data sets. Using these minimization criteria result in the following values of $N_c$ for the German, Australian and Japanese data sets
\begin{equation*}
 N_c^\text{Germany} = 6, \quad N_c^\text{Australia} = 10, \quad N_c^\text{Japan} = 9,
\end{equation*}
respectively. Next using the methodology in section~\ref{sec:methodprob}, we label all the clusters with either 0 or 1. Figure~\ref{fig:labeling_clusters}, panel (a) shows $6$ components for the German data set. Below the graph, a table shows the number of 0s and 1s in each cluster. All the clusters are then labeled according to 2.5.1, equation~\eqref{eq:labeling_cluster}. The train set results of Australian and Japanese data are displayed in panels (b) and (c), respectively. The classification accuracy in the train set is 77.40\%, 83.91\% and 82.27\% for German, Australian and Japanese train sets, respectively.

Next, we use the labeled clusters to label individual testing data according to 2.5.4, equation~\eqref{eq:labeling_data} with decision boundary $D = 0.5$. The corresponding results with tabulated number of 0s and 1s are displayed in figure~\ref{fig:labeling_clusters}, panels (d), (e) and (f) for German, Australian, and Japanese test sets, respectively. Test set Accuracy for German, Australian and Japanese data is reported as 70.35\%, 84.78\% and 83.71\%, respectively. Table \ref{class_acc} displays the class accuracy for both train and test sets. The blank fields show that no test samples belong to the corresponding class. 

We have an interesting observation in figure~\ref{fig:labeling_clusters}, panels (a) and (d). In panel (a), the third cluster shows train accuracy of $99.02\%$ as most of the assigned members to this cluster are labeled 0. However, according to test set in panel (d), there are only 7 members with label 0 and the accuracy of this cluster is $46.66\%$. This is due to the fact that in the train set, we have employed SMOTE to increase the number of minority class with label 0, whereas in the test set, the data is imbalanced.

It is noticeable that features do not necessarily follow the normal distribution and any non-normality in the feature space is captured by GMM. Increasing the number of components beyond two leads to higher accuracy of the algorithm. 

The confusion matrices
\begin{equation}
 C_\text{G}=\begin{pmatrix}
307&151\\
56&402\\
\end{pmatrix},
C_\text{A}=\begin{pmatrix}
203&51\\
23&183\\
\end{pmatrix}, 
C_\text{J}=\begin{pmatrix}
220&14\\
64&142\\
\end{pmatrix} , \nonumber
\end{equation}
are derived from German, Australian and Japanese train sets, respectively. Their test set counterparts are \begin{equation}
C_\text{G}=\begin{pmatrix}
9&83\\
16&226\\
\end{pmatrix},
C_\text{A}=\begin{pmatrix}
104&25\\
10&91\\
\end{pmatrix}, C_\text{J}=\begin{pmatrix}
125&5\\
31&60\\
\end{pmatrix} . \nonumber
\end{equation}
The first column shows the number of true negatives (TN) and false negatives (FN), and the second column shows the number of false positives (FP) and true positives (TP), respectively.\footnote{see \cite{geron2017hands} and Scikit-learn package \cite{pedregosa2011scikit}
}

\begin{figure*}
 \centering
 \subfigure[Australia]{{\includegraphics[width=4.8cm]{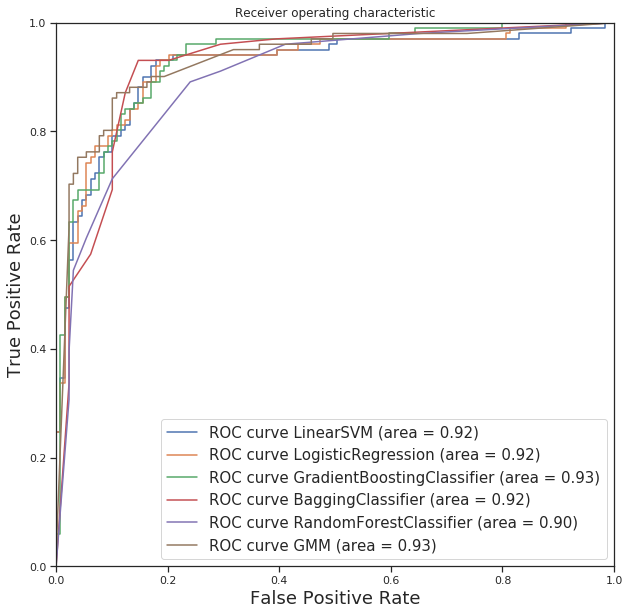}}}
 \quad
 \subfigure[Japan]{{\includegraphics[width=4.8cm]{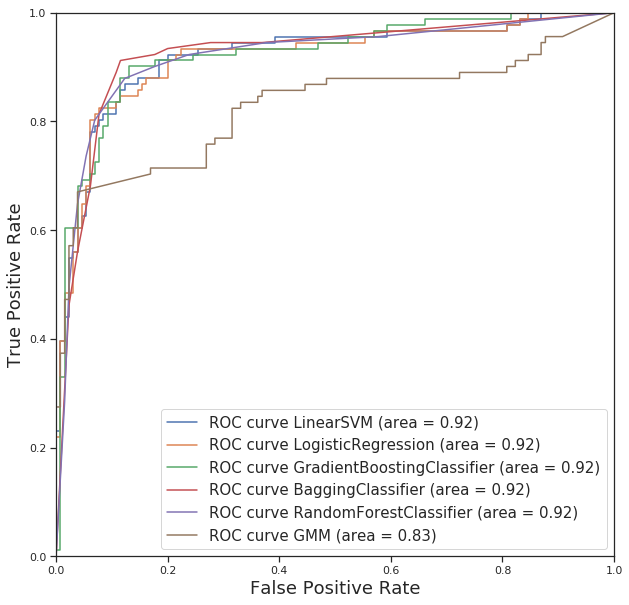} }}%
 \quad
 \subfigure[German]{{\includegraphics[width=4.8cm]{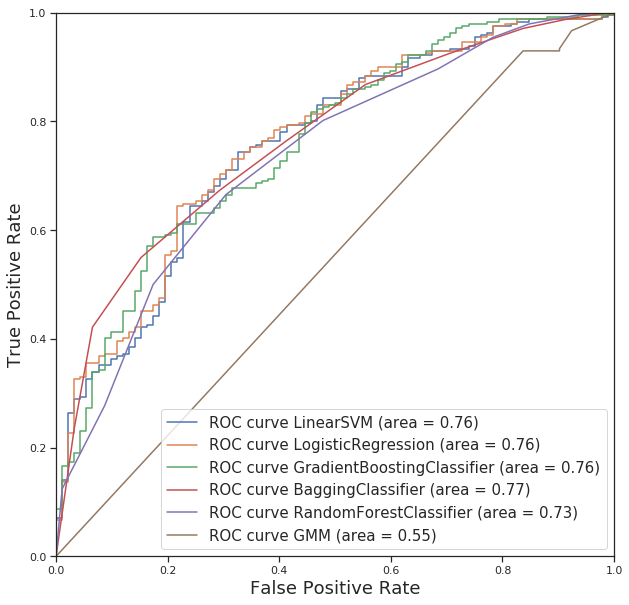} }}%
 \caption{ROC curves for Australian, Japanese and German data sets with respect to different classification methods. }%
 \label{ROC}%
\end{figure*}

\subsection{Benchmark Experiment}

This section includes a comparison of our proposed model with other traditional credit scoring models, namely Random Forest Classifier (RFC), Bagging Classifier (BC), Gradient Boosting Classifier (GBC), Logistic Regression (LR) and Linear Support Vector Machine (LSVM). We used the same train and test data for making a comparison between models. 
The accuracy of different models are shown in table~\ref{tab:comparison}. 
Our CSGM model performance on the test set is fairly acceptable compared to others. Traditional models such as Bagging Classifier and Random Forest Classifier show train accuracy of as high as $100\%$, Whereas their test set accuracy drops to as low as $71\%$. Fortunately, unlike other methods, CSGM model without using cross validation techniques does not exhibit over-fitting. 
For making a more precise comparison between CSGM model versus others, four accuracy measures are used, namely Recall score, Precision score, $F_1$ score and AUC score.

Recall Score and Precision score are defined as $\frac{TP}{TP+FN}$ and $\frac{TP}{TP+FP}$, respectively. $F_1$ score combines the two mentioned accuracy measures as
\begin{equation}
 F_1 = 2\left(\frac{1}{\text{Precision}}+\frac{1}{\text{Recall}}\right)^{-1} ,
\end{equation}
which gives more weight to lower of Recall and Precision scores. The ROC curves for different methods are shown in figure~\ref{ROC}. ROC curves plot the true positive rates against false positive rates for different thresholds. AUC score measures the area under ROC curve. 
Table~\ref{tab:comparison} shows that for the German test set, our CSGM model performs best compared to other models with respect to Recall score. For Japanese and Australian test sets, CSGM model outperforms others by Precision score and AUC score measurements, respectively.
In addition, for the Australian test set, CSGM model performs almost as well as LSVC and better than others using $F_1$ score, Recall and AUC score measurements. For Japanese test set, CSGM model works better using Precision score, which is most suited for banks with higher risk appetite which are willing to take more risk by issuing loans to riskier costumers. The AUC score for German train and test set suggest that when dealing with imbalanced data, CSGM model does not perform well in test set.

\begin{table}
  \centering
  \caption{Comparison between training and testing accuracy of different models}
\resizebox{\textwidth}{!}{
\begin{tabular}{lccc}
\multicolumn{4}{c}{Train Set}\\
    \toprule
    Model & Germany & Australia & Japan \\
    \midrule
    LSVM  & 85.26\% & 86.09\% & 87.95\% \\
    \textcolor[rgb]{ .129,  .129,  .129}{LR} & 84.83\% & 87.17\% & 88.86\% \\
    \textcolor[rgb]{ .129,  .129,  .129}{GBCLF} & 94.21\% & 98.04\% & 99.32\% \\
    \textcolor[rgb]{ .129,  .129,  .129}{BCLF} & 99.56\% & 98.70\% & 97.73\% \\
    \textcolor[rgb]{ .129,  .129,  .129}{RF} & 99.89\% & 99.35\% & 98.18\% \\
    GMM   & 77.40\% & 83.91\% & 82.27\% \\
    \bottomrule
    \end{tabular}%
    \qquad 
\begin{tabular}{lccc}
\multicolumn{4}{c}{Test Set}\\
    \toprule
    Model & \multicolumn{1}{c}{Germany} & \multicolumn{1}{c}{Australia} & \multicolumn{1}{c}{Japan} \\
    \midrule
    LSVM  & 75.15\% & 86.52\% & 84.62\% \\
    \textcolor[rgb]{ .129,  .129,  .129}{LR} & 74.85\% & 84.78\% & 85.52\% \\
    \textcolor[rgb]{ .129,  .129,  .129}{GBCLF} & 74.85\% & 84.35\% & 87.33\% \\
    \textcolor[rgb]{ .129,  .129,  .129}{BCLF} & 71.56\% & 83.91\% & 89.14\% \\
    \textcolor[rgb]{ .129,  .129,  .129}{RF} & 72.46\% & 81.74\% & 87.78\% \\
    GMM   & 70.35\% & 84.78\% & 83.71\% \\
    \bottomrule
    \end{tabular}%
}
\label{tab:comparisontest}%
\end{table}%

\begin{table}
\centering
\caption{Comparison between different models in training and testing set of German, Japanese and Australian data.}
\resizebox{\textwidth}{!}{
\begin{tabular}{lcccccccccccccc}
\toprule
\multirow{4}[5]{*}{Models} & \multicolumn{14}{c}{Training Results} \\
\cmidrule{2-15}      & \multicolumn{4}{c}{Germany}   & $\qquad$ & \multicolumn{4}{c}{Japan}     & $\qquad$ & \multicolumn{4}{c}{Australia} \\
\cmidrule{2-15}      & AUC  & precision  & recall  & F1    & $\qquad$ & AUC  & precision  & recall  & F1    & $\qquad$ & AUC  & precision  & recall  & F1  \\
      & score & score & score & score & $\qquad$ & score & score & score & score & $\qquad$ & score & score & score & score \\
\midrule
Linear SVM & 93\%  & 83\%  & 88\%  & 86\%  & $\qquad$ & 94\%  & 82\%  & 95\%  & 88\%  & $\qquad$ & 93\%  & 80\%  & 92\%  & 86\% \\
Logistic Regression & 93\%  & 83\%  & 87\%  & 85\%  & $\qquad$ & 94\%  & 85\%  & 93\%  & 89\%  & $\qquad$ & 93\%  & 84\%  & 88\%  & 86\% \\
Gradient Boosting Classifier & 99\%  & 93\%  & 95\%  & 94\%  & $\qquad$ & 100\% & 99\%  & 100\% & 99\%  & $\qquad$ & 100\% & 99\%  & 97\%  & 98\% \\
Bagging Classifier & 100\% & 100\% & 99\%  & 100\% & $\qquad$ & 100\% & 99\%  & 96\%  & 98\%  & $\qquad$ & 100\% & 100\% & 97\%  & 99\% \\
Random Forest Classifier & 100\% & 100\% & 99\%  & 100\% & $\qquad$ & 100\% & 99\%  & 96\%  & 98\%  & $\qquad$ & 100\% & 100\% & 97\%  & 99\% \\
Gaussian Mixture Model & 81\%  & 73\%  & 88\%  & 80\%  & $\qquad$ & 86\%  & 91\%  & 69\%  & 78\%  & $\qquad$ & 89\%  & 78\%  & 89\%  & 83\% \\
\midrule
\multirow{4}[5]{*}{Models} & \multicolumn{14}{c}{Testing Results} \\
\cmidrule{2-15}      & \multicolumn{4}{c}{Germany}   & $\qquad$ & \multicolumn{4}{c}{Japan}     & $\qquad$ & \multicolumn{4}{c}{Australia} \\
\cmidrule{2-15}      & AUC  & Precision  & Recall  & F1    & $\qquad$ & AUC  & Precision  & Recall  & F1    & $\qquad$ & AUC  & Precision  & Recall  & F1  \\
      & Score & Score & Score & Score & $\qquad$ & Score & Score & Score & Score & $\qquad$ & Score & Score & Score & Score \\
Linear SVM & 76\%  & 81\%  & 86\%  & 83\%  & $\qquad$ & 92\%  & 77\%  & 89\%  & 83\%  & $\qquad$ & 92\%  & 80\%  & 92\%  & 86\% \\
Logistic Regression & 76\%  & 81\%  & 85\%  & 83\%  & $\qquad$ & 92\%  & 79\%  & 88\%  & 83\%  & $\qquad$ & 92\%  & 82\%  & 84\%  & 83\% \\
Gradient Boosting Classifier & 76\%  & 81\%  & 86\%  & 83\%  & $\qquad$ & 92\%  & 84\%  & 86\%  & 85\%  & $\qquad$ & 93\%  & 85\%  & 78\%  & 81\% \\
Bagging Classifier & 77\%  & 83\%  & 76\%  & 80\%  & $\qquad$ & 92\%  & 85\%  & 89\%  & 87\%  & $\qquad$ & 92\%  & 86\%  & 76\%  & 81\% \\
Random Forest Classifier & 73\%  & 82\%  & 80\%  & 81\%  & $\qquad$ & 92\%  & 89\%  & 80\%  & 84\%  & $\qquad$ & 90\%  & 85\%  & 71\%  & 77\% \\
Gaussian Mixture Model & 55\%  & 73\%  & 93\%  & 82\%  & $\qquad$ & 83\%  & 92\%  & 66\%  & 77\%  & $\qquad$ & 93\%  & 78\%  & 90\%  & 84\% \\
\bottomrule
\end{tabular}%
}
\label{tab:comparison}%
\end{table}%
\section{Executive Insights}\label{sec:business}

\begin{figure*}
    \centering
    \subfigure[US Consumer Loans in Billions of Dollars]{{\includegraphics[width=7.0cm]{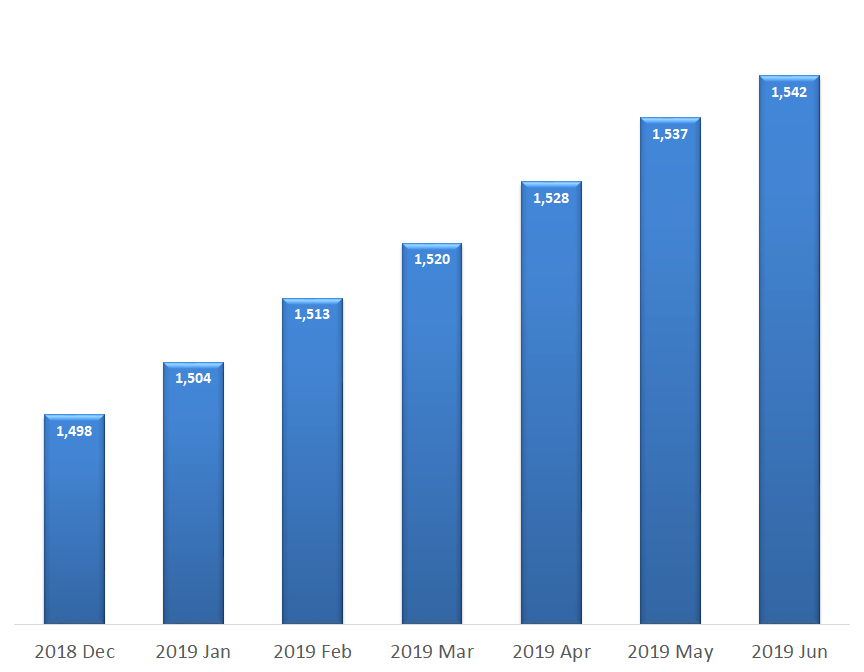} }}%
    \quad
    \subfigure[2016 Credit Card payments in Billions of Dollars]{{\includegraphics[width=7.0cm]{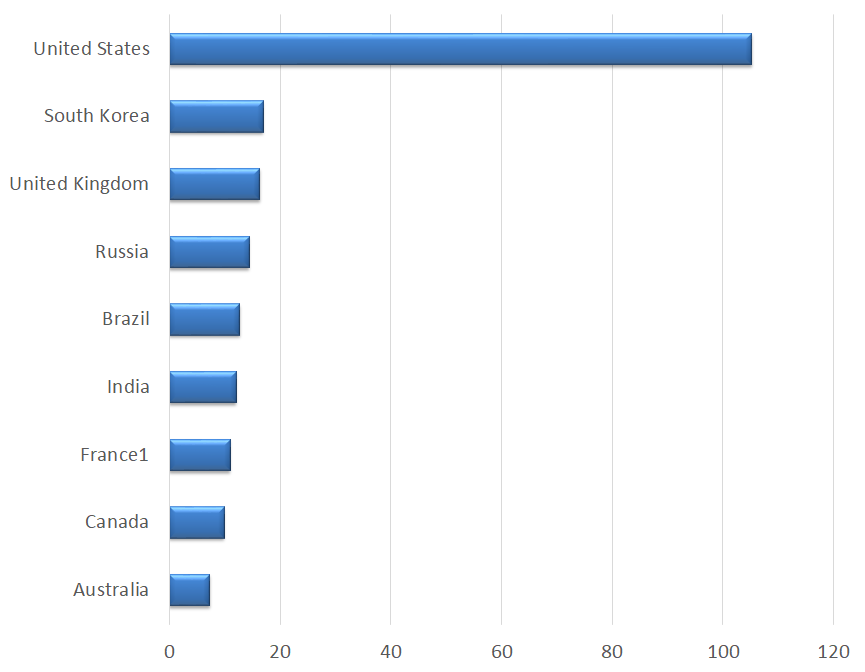} }}%
    \caption{Consumer Credit Scope in the United States and Worldwide}%
    \label{fig:creditscoringscope}%
\end{figure*}

Intelligent predictive models have gained considerable attention for credit risk assessment as a result of the latest sub-prime crisis and growing complexity of world economy. During the sub-prime crisis, had the business executives and other finance professionals had access to more efficient predictive models for credit risk measurement, they could have avoided such substantial losses. In terms of business impact, credit scoring scope in the financial industry is quite significant. For instance, in June 2019, the consumer loans held by US citizens were reported as 1.542 trillion dollars. In figure~\ref{fig:creditscoringscope}, panel (a), consumer loans from December 2018 to June 2019 in Billions of dollars is reported by Fed (Federal Reserve Statistical Release, July 12, 2019, Assets and Liabilities of Commercial Banks in the United States). In panel (b), 2016 Credit Card payments in Billions of Dollars is shown (BIS, Committee on Payments and Market Infrastructures, Statistics on payment, clearing and settlement systems in the CPMI countries, December 2017, page 429 table 7). As reported by~\cite{khandani2010consumer}, the cost saving for credit lenders by using Machine Learning applications in their business is quite substantial and ranges from $6\%$ to $25\%$.

Probabilistic approaches for classification are strong tools for credit scoring purposes. One of the main CSGM's benefits is that each forecast takes the form of a predictive distribution rather than a single point estimate. Flexibility of the probabilistic classification algorithms can be essential for decision making duties of risk managers in credit lending industry. This is due to the fact that managers can make a tradeoff between precision and recall by adjusting the decision boundary. A classifier that dismisses too many creditworthy costumers and accepts only safest ones would have a low recall and a high precision score. On the other way around, a classifier which keeps many costumers with bad credit, has a high recall and a low precision score. Fortunately, CSGM model is flexible enough to provide both functionalities in a unified classifier. By setting an appropriate decision boundary, CSGM would be able to set a proper balance between recall and precision in accordance with risk tolerance of a financial manager. 

CSGM model provides a computationally efficient, powerful, flexible tool for decision making. CSGM model's learning time is quite short and financial institutions can save a lot of running time to build the model even when the data set is large. The power of Gaussian Mixture models to recognize non-convex cluster patterns with arbitrary shapes from the training data using Expectation-Maximization algorithm is another powerful benefit of the model. CSGM model not only points out which customers are more likely to default, but also implies similarities between new customers and old customers when it includes them into same Gaussian clusters. This have useful implications for decision makers in the credit lending business, as we know much about how old customers behave. Compared to other traditional models, CSGM is superior because it can provide valuable information for senior management and associated decision makers by putting loan applicants into meaningful categories. Our model can categorize customers into groups according to similarities between their high dimensional features using latent variables. Latent variables in CSGM model therefore make possible to view what would be impossible to observe by a busy manager's naked eye.

\section{Conclusions and Future Research}\label{sec:con}

CSGM model compared to other approaches such as Support Vector Machines and Decision Trees, provides a different modeling framework and mathematical structure. In this paper, we empirically tested five models with our CSGM approach on three real world credit scoring data. The CSGM classifier presented in our study, showed superior capability compared to other models in term of over-fitting and under-fitting. This shows the model's flexibility to create a tradeoff between bias and variance. Almost all models were generally over-fitted across all three data-sets. However, the CSGM classifier shows high performance in most test cases and does not exhibit over-fitting. 

Ensemble classification which employs popular methods such as Bagging, Random Subspace Method (RSM), and Boosting, can be used along with CSGM model to increase the accuracy and minimize over-fitting of our credit scoring model. Even though, as shown in Section~\ref{sec:emp}, our model's performance on train and test sets are already comparable with other models, using methods such as cross validation techniques might lead to improvements in model's performance.

We plan to apply the current classification framework as a future line of research and extend its capabilities to tackle other problems in more unstable and uncertain modeling environments. We believe that probabilistic classification such as CSGM will prove more efficient when we are faced with higher degrees of uncertainty. On the modeling side, trying mixtures of other probability measures such as student's t-distribution would be interesting. This would be exciting to see if the powerful feature of variance and bias balance which CSGM model possesses is intact when normal component distributions are replaced with other types of distributions. We also would like to apply the same approach to more sophisticated and populated data sets with higher number of features and see if probabilistic classification advantages are more evident in a larger testing environment. 

\clearpage

\setcitestyle{numbers} 
\bibliographystyle{plainnat}
\bibliography{refs}

\begin{thebibliography}{36}
\providecommand{\natexlab}[1]{#1}
\providecommand{\url}[1]{\texttt{#1}}
\expandafter\ifx\csname urlstyle\endcsname\relax
  \providecommand{\doi}[1]{doi: #1}\else
  \providecommand{\doi}{doi: \begingroup \urlstyle{rm}\Url}\fi

\bibitem[Akaike(1974)]{akaike1974new}
Hirotugu Akaike.
\newblock A new look at the statistical model identification.
\newblock In \emph{Selected Papers of Hirotugu Akaike}, pages 215--222.
  Springer, 1974.

\bibitem[Baesens et~al.(2003)Baesens, Van~Gestel, Viaene, Stepanova, Suykens,
  and Vanthienen]{baesens2003benchmarking}
Bart Baesens, Tony Van~Gestel, Stijn Viaene, Maria Stepanova, Johan Suykens,
  and Jan Vanthienen.
\newblock Benchmarking state-of-the-art classification algorithms for credit
  scoring.
\newblock \emph{Journal of the operational research society}, 54\penalty0
  (6):\penalty0 627--635, 2003.

\bibitem[Bishop(2006)]{bishop2006pattern}
Christopher~M Bishop.
\newblock \emph{Pattern recognition and machine learning}.
\newblock springer, 2006.

\bibitem[Brown and Mues(2012)]{brown2012experimental}
Iain Brown and Christophe Mues.
\newblock An experimental comparison of classification algorithms for
  imbalanced credit scoring data sets.
\newblock \emph{Expert Systems with Applications}, 39\penalty0 (3):\penalty0
  3446--3453, 2012.

\bibitem[Chawla et~al.(2002)Chawla, Bowyer, Hall, and
  Kegelmeyer]{chawla2002smote}
Nitesh~V Chawla, Kevin~W Bowyer, Lawrence~O Hall, and W~Philip Kegelmeyer.
\newblock Smote: synthetic minority over-sampling technique.
\newblock \emph{Journal of artificial intelligence research}, 16:\penalty0
  321--357, 2002.

\bibitem[Chen and Li(2010)]{chen2010combination}
Fei-Long Chen and Feng-Chia Li.
\newblock Combination of feature selection approaches with svm in credit
  scoring.
\newblock \emph{Expert systems with applications}, 37\penalty0 (7):\penalty0
  4902--4909, 2010.

\bibitem[Chen et~al.(2012)Chen, Xiang, Liu, and Wang]{chen2012credit}
Weimin Chen, Guocheng Xiang, Youjin Liu, and Kexi Wang.
\newblock Credit risk evaluation by hybrid data mining technique.
\newblock \emph{Systems Engineering Procedia}, 3:\penalty0 194--200, 2012.

\bibitem[Dempster et~al.(1977)Dempster, Laird, and Rubin]{dempster1977maximum}
Arthur~P Dempster, Nan~M Laird, and Donald~B Rubin.
\newblock Maximum likelihood from incomplete data via the em algorithm.
\newblock \emph{Journal of the Royal Statistical Society: Series B
  (Methodological)}, 39\penalty0 (1):\penalty0 1--22, 1977.

\bibitem[Dua and Graff(2017)]{Dua:2019}
Dheeru Dua and Casey Graff.
\newblock {UCI} machine learning repository, 2017.
\newblock URL \url{http://archive.ics.uci.edu/ml}.

\bibitem[G{\'e}ron(2017)]{geron2017hands}
Aur{\'e}lien G{\'e}ron.
\newblock \emph{Hands-on machine learning with Scikit-Learn and TensorFlow:
  concepts, tools, and techniques to build intelligent systems}.
\newblock " O'Reilly Media, Inc.", 2017.

\bibitem[Ghodselahi(2011)]{ghodselahi2011hybrid}
Ahmad Ghodselahi.
\newblock A hybrid support vector machine ensemble model for credit scoring.
\newblock \emph{International Journal of Computer Applications}, 17\penalty0
  (5):\penalty0 1--5, 2011.

\bibitem[Han et~al.(2006)Han, Kamber, and Pei]{han2006data}
Jiawei Han, Micheline Kamber, and Jian Pei.
\newblock Data mining: concepts and techniques. 2001.
\newblock \emph{San Francisco: Morgan Kauffman}, 2006.

\bibitem[Han et~al.(2019)Han, Cui, Lan, Kang, Deng, and Jia]{han2019gaussian}
Xu~Han, Runbang Cui, Yanfei Lan, Yanzhe Kang, Jiang Deng, and Ning Jia.
\newblock A gaussian mixture model based combined resampling algorithm for
  classification of imbalanced credit data sets.
\newblock \emph{International Journal of Machine Learning and Cybernetics},
  pages 1--13, 2019.

\bibitem[Hand and Henley(1997)]{hand1997statistical}
David~J Hand and William~E Henley.
\newblock Statistical classification methods in consumer credit scoring: a
  review.
\newblock \emph{Journal of the Royal Statistical Society: Series A (Statistics
  in Society)}, 160\penalty0 (3):\penalty0 523--541, 1997.

\bibitem[Harris(2015)]{harris2015credit}
Terry Harris.
\newblock Credit scoring using the clustered support vector machine.
\newblock \emph{Expert Systems with Applications}, 42\penalty0 (2):\penalty0
  741--750, 2015.

\bibitem[Hoffmann et~al.(2002)Hoffmann, Baesens, Martens, Put, and
  Vanthienen]{hoffmann2002comparing}
Frank Hoffmann, Bart Baesens, Jurgen Martens, Ferdi Put, and Jan Vanthienen.
\newblock Comparing a genetic fuzzy and a neurofuzzy classifier for credit
  scoring.
\newblock \emph{International Journal of Intelligent Systems}, 17\penalty0
  (11):\penalty0 1067--1083, 2002.

\bibitem[Hsieh(2005)]{hsieh2005hybrid}
Nan-Chen Hsieh.
\newblock Hybrid mining approach in the design of credit scoring models.
\newblock \emph{Expert systems with applications}, 28\penalty0 (4):\penalty0
  655--665, 2005.

\bibitem[Hsieh and Hung(2010)]{hsieh2010data}
Nan-Chen Hsieh and Lun-Ping Hung.
\newblock A data driven ensemble classifier for credit scoring analysis.
\newblock \emph{Expert systems with Applications}, 37\penalty0 (1):\penalty0
  534--545, 2010.

\bibitem[Khandani et~al.(2010)Khandani, Kim, and Lo]{khandani2010consumer}
Amir~E Khandani, Adlar~J Kim, and Andrew~W Lo.
\newblock Consumer credit-risk models via machine-learning algorithms.
\newblock \emph{Journal of Banking \& Finance}, 34\penalty0 (11):\penalty0
  2767--2787, 2010.

\bibitem[Lee and Chen(2005)]{lee2005two}
Tian-Shyug Lee and I-Fei Chen.
\newblock A two-stage hybrid credit scoring model using artificial neural
  networks and multivariate adaptive regression splines.
\newblock \emph{Expert Systems with Applications}, 28\penalty0 (4):\penalty0
  743--752, 2005.

\bibitem[Lessmann et~al.(2015)Lessmann, Baesens, Seow, and
  Thomas]{lessmann2015benchmarking}
Stefan Lessmann, Bart Baesens, Hsin-Vonn Seow, and Lyn~C Thomas.
\newblock Benchmarking state-of-the-art classification algorithms for credit
  scoring: An update of research.
\newblock \emph{European Journal of Operational Research}, 247\penalty0
  (1):\penalty0 124--136, 2015.

\bibitem[Liu and Schumann(2005)]{liu2005data}
Y~Liu and M~Schumann.
\newblock Data mining feature selection for credit scoring models.
\newblock \emph{Journal of the Operational Research Society}, 56\penalty0
  (9):\penalty0 1099--1108, 2005.

\bibitem[Louzada et~al.(2016)Louzada, Ara, and
  Fernandes]{louzada2016classification}
Francisco Louzada, Anderson Ara, and Guilherme~B Fernandes.
\newblock Classification methods applied to credit scoring: Systematic review
  and overall comparison.
\newblock \emph{Surveys in Operations Research and Management Science},
  21\penalty0 (2):\penalty0 117--134, 2016.

\bibitem[Martens et~al.(2007)Martens, Baesens, Van~Gestel, and
  Vanthienen]{martens2007comprehensible}
David Martens, Bart Baesens, Tony Van~Gestel, and Jan Vanthienen.
\newblock Comprehensible credit scoring models using rule extraction from
  support vector machines.
\newblock \emph{European journal of operational research}, 183\penalty0
  (3):\penalty0 1466--1476, 2007.

\bibitem[Nanni and Lumini(2009)]{nanni2009experimental}
Loris Nanni and Alessandra Lumini.
\newblock An experimental comparison of ensemble of classifiers for bankruptcy
  prediction and credit scoring.
\newblock \emph{Expert systems with applications}, 36\penalty0 (2):\penalty0
  3028--3033, 2009.

\bibitem[Paleologo et~al.(2010)Paleologo, Elisseeff, and
  Antonini]{paleologo2010subagging}
Giuseppe Paleologo, Andr{\'e} Elisseeff, and Gianluca Antonini.
\newblock Subagging for credit scoring models.
\newblock \emph{European Journal of Operational Research}, 201\penalty0
  (2):\penalty0 490--499, 2010.

\bibitem[Pedregosa et~al.(2011)Pedregosa, Varoquaux, Gramfort, Michel, Thirion,
  Grisel, Blondel, Prettenhofer, Weiss, Dubourg, et~al.]{pedregosa2011scikit}
Fabian Pedregosa, Ga{\"e}l Varoquaux, Alexandre Gramfort, Vincent Michel,
  Bertrand Thirion, Olivier Grisel, Mathieu Blondel, Peter Prettenhofer, Ron
  Weiss, Vincent Dubourg, et~al.
\newblock Scikit-learn: Machine learning in python.
\newblock \emph{Journal of machine learning research}, 12\penalty0
  (Oct):\penalty0 2825--2830, 2011.

\bibitem[Schwarz et~al.(1978)]{schwarz1978estimating}
Gideon Schwarz et~al.
\newblock Estimating the dimension of a model.
\newblock \emph{The annals of statistics}, 6\penalty0 (2):\penalty0 461--464,
  1978.

\bibitem[Thomas(2000)]{thomas2000survey}
Lyn~C Thomas.
\newblock A survey of credit and behavioural scoring: forecasting financial
  risk of lending to consumers.
\newblock \emph{International journal of forecasting}, 16\penalty0
  (2):\penalty0 149--172, 2000.

\bibitem[Thomas et~al.(2002)Thomas, Edelman, and Crook]{thomas2002credit}
Lyn~C Thomas, David~B Edelman, and Jonathan~N Crook.
\newblock \emph{Credit scoring and its applications}.
\newblock SIAM, 2002.

\bibitem[Tsai and Wu(2008)]{tsai2008using}
Chih-Fong Tsai and Jhen-Wei Wu.
\newblock Using neural network ensembles for bankruptcy prediction and credit
  scoring.
\newblock \emph{Expert systems with applications}, 34\penalty0 (4):\penalty0
  2639--2649, 2008.

\bibitem[Van~Gestel et~al.(2006)Van~Gestel, Baesens, Suykens, Van~den Poel,
  Baestaens, and Willekens]{van2006bayesian}
Tony Van~Gestel, Bart Baesens, Johan~AK Suykens, Dirk Van~den Poel, Dirk-Emma
  Baestaens, and Marleen Willekens.
\newblock Bayesian kernel based classification for financial distress
  detection.
\newblock \emph{European journal of operational research}, 172\penalty0
  (3):\penalty0 979--1003, 2006.

\bibitem[Wang et~al.(2011)Wang, Hao, Ma, and Jiang]{wang2011comparative}
Gang Wang, Jinxing Hao, Jian Ma, and Hongbing Jiang.
\newblock A comparative assessment of ensemble learning for credit scoring.
\newblock \emph{Expert systems with applications}, 38\penalty0 (1):\penalty0
  223--230, 2011.

\bibitem[West(2000)]{west2000neural}
David West.
\newblock Neural network credit scoring models.
\newblock \emph{Computers \& Operations Research}, 27\penalty0
  (11-12):\penalty0 1131--1152, 2000.

\bibitem[Xiao et~al.(2016)Xiao, Xiao, and Wang]{xiao2016ensemble}
Hongshan Xiao, Zhi Xiao, and Yu~Wang.
\newblock Ensemble classification based on supervised clustering for credit
  scoring.
\newblock \emph{Applied Soft Computing}, 43:\penalty0 73--86, 2016.

\bibitem[Zhang et~al.(2010)Zhang, Zhou, Leung, and Zheng]{zhang2010vertical}
Defu Zhang, Xiyue Zhou, Stephen~CH Leung, and Jiemin Zheng.
\newblock Vertical bagging decision trees model for credit scoring.
\newblock \emph{Expert Systems with Applications}, 37\penalty0 (12):\penalty0
  7838--7843, 2010.

\end{thebibliography}



\end{document}